\documentclass[letterpaper, 10 pt, conference]{ieeeconf}  % Comment this line out if you need a4paper

\IEEEoverridecommandlockouts                              % This command is only needed if 
\usepackage{graphics} % for pdf, bitmapped graphics files
\usepackage{algorithm}
\usepackage{algpseudocode}

\usepackage{xcolor}
\usepackage{soul}

\title{\LARGE \bf
Toward the use of proxies for efficient learning manipulation and locomotion strategies on soft robots.

}

\author{Etienne Ménager$^{*}$, Quentin Peyron$^{*}$ and Christian Duriez$^{*}$% <-this % stops a space
\thanks{$^{*}$Univ. Lille, Inria, CNRS, Centrale Lille, UMR 9189 CRIStAL, F-59000 Lille, France
        {\tt\small etienne.menager@inria.fr}}%
}

\begin{document}

\maketitle
\thispagestyle{empty}
\pagestyle{empty}

%%%%%%%%%%%%%%%%%%%%%%%%%%%%%%%%%%%%%%%%%%%%%%%%%%%%%%%%%%%%%%%%%%%%%%%%%%%%%%%%
\begin{abstract}

Soft robots are naturally designed to perform safe interactions with their environment, like locomotion and manipulation. In the literature, there are now many concepts, often bio-inspired, to propose new modes of locomotion or grasping. However, a methodology for implementing motion planning of these tasks, as exists for rigid robots, is still lacking. One of the difficulties comes from the modeling of these robots, which is very different, as it is based on the mechanics of deformable bodies. These models, whose dimension is often very large, make learning and optimization methods very costly. In this paper, we propose a proxy approach, as exists for humanoid robotics. This proxy is a simplified model of the robot that enables frugal learning of a motion strategy. This strategy is then transferred to the complete model to obtain the corresponding actuation inputs. Our methodology is illustrated and analyzed on two classical designs of soft robots doing manipulation and locomotion tasks.

\end{abstract}


%%%%%%%%%%%%%%%%%%%%%%%%%%%%%%%%%%%%%%%%%%%%%%%%%%%%%%%%%%%%%%%%%%%%%%%%%%%%%%%%
\section{Introduction}
\label{sec:introduction}

Robots made of soft materials have intrinsic flexibility and adaptability to their environment \cite{c1}. However, the high compliance of soft robots can lead to challenges in their control when performing complex tasks such as grasping and manipulation of an object or walking efficiently in a given direction. Solving these tasks, which we call \emph{Tasks with Sequential Configurations and Contacts} (TSCC) in this paper, require to move the robot in a succession of configurations which exploit numerous and varying contacts with the environment (ground, object). Even though open loop and closed loop control can be derived from numerical models \cite{c2}, planning the target configurations to reach is generally difficult to do. TSCC can be solved using motion planning strategies based on learning algorithms, including Reinforcement Learning (RL) algorithms. The use of RL algorithms in soft-robotics presents specific challenges \cite{c3, c9}, but can lead to the solution of manipulation \cite{c4} or locomotion \cite{c5} problems. Other learning approaches, such as supervised learning, or model-based and model-free RL variations have been used for soft-robotic control \cite{c21}.

Computer simulations provide efficient data for learning algorithms \cite{c2a, c2b, c19, c20}. The simulators use different calculation methods to simulate soft robots, which choice involves a trade-off between accuracy and calculation time. Some methods focus on calculation time, such as pseudo-rigid bodies \cite{c2b}, geometric approaches \cite{c21}, and particles \cite{c19}, while others focus on modeling physical laws such as the Finite Element Method (FEM) \cite{c3}. These different simulators have been used to collect data for imitation learning or pre-training methods \cite{c9, c22} and to implement classical path planning algorithms such as the rapidly-exploring random tree (RRT) \cite{c23, c24, c25}. 

However, these motion planning methods either lack genericity or are cumbersome to compute due to the model they use. The pseudo-rigid bodies methods use parameters without physical meaning. The particles methods lack the ability to simulate a wide range of geometry and actuation. RTT has been implemented with geometric models whose application is limited to slender soft robots and positioning tasks \cite{c25}. The extension to more complex geometries and contact configuration is not trivial. FEM are good candidates to handle in a generic way complex soft robots with non-regular geometry, soft actuators, and actuation redundancy, for which they have been validated experimentally \cite{c2, c16}. However, the presence of many contacts and constraints in the environments in TSCC can lead to demanding computation times \cite{c3}. As a consequence, there is no general and efficient motion planning method for complex soft robots performing these tasks.

In this work, we make the assumption that FEM provides a faithful model of the robot and we look for a computationally efficient motion planning algorithm. The approaches developed in rigid robotics are not directly applicable but can be a source of inspiration. A humanoid robot is a good example of a complex robotic system with many degrees of freedom and redundancy of actuation. For this kind of system, the motion planning problem is divided in two parts: the generation of a feasible trajectory in the configuration space and the transfer of this trajectory to the actuation space. The first step is done using simplified models of the robots, called proxy models \cite{c26, c28, c11}. This allows to simulate a simplified dynamic model at each time step during the motion planning phase. This trajectory is then used as a reference to control the movement of the complete robot in simulation. This allows to generate the corresponding trajectory in the actuation space, dealing with eventual actuation redundancy. The idea is to apply a similar method to soft robots, solving part of the motion generation problem with a simplified model.

\begin{figure}
\centering
\resizebox{.45\textwidth}{!}{
\includegraphics{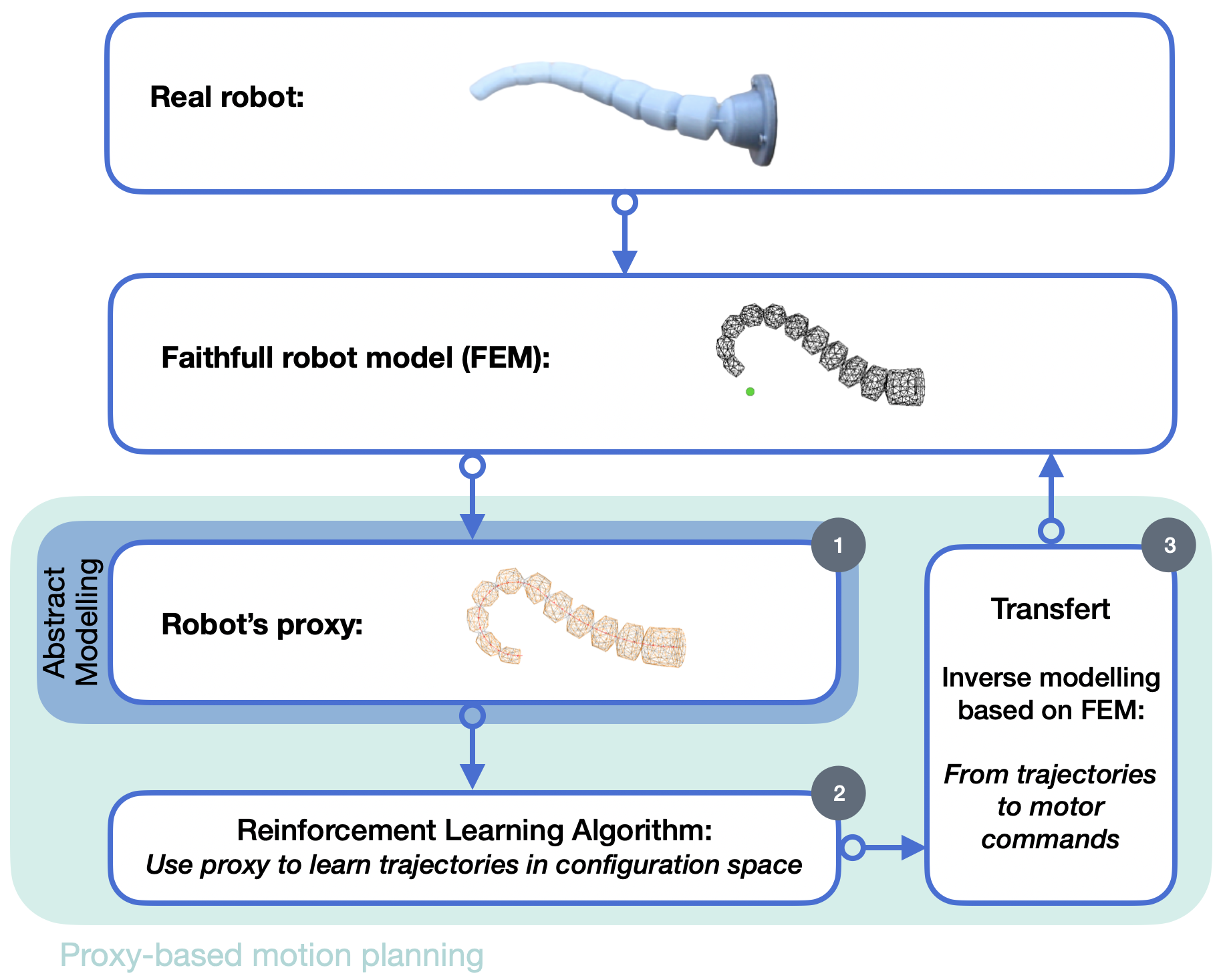}
}
\caption{Pipeline for using the proxy model to solve a given task in the case of the trunk like robot. Starting from a complete model, an abstract model called proxy is created (1). The proxy is then used to perform the learning and find a realistic trajectory of the robot (2). The actual actuation of the robot is calculated from the learned trajectory using an inverse model (3). The obtained motor control is applied to the complete model of the robot. }
\label{im:idea}
\end{figure}

\textit{Contributions}: In this paper, assuming we have a high fidelity FEM model of the soft robot, we propose a new motion planning to perform locomotion and manipulation tasks based on computer simulation. The method is composed of 3 steps, as illustrated in Figure 1. The first step is to create a proxy model of the robot, i.e. to extract from the faithful model a model less complex to simulate. At step 2, the proxy is used to generate a trajectory in the configuration space with a learning approach. During step 3, the successive positions of the generated trajectory are used as a reference for a position controller based on the inverse FEM model. We will see that in addition to being a frugal method, i.e. requiring less resources and computation time, the proxy model facilitates learning with smaller or equal action space dimension than the number of actuators and an observation space directly defined by the model. We experiment this pipeline with two different robots that have to perform respectively a locomotion task and a manipulation task. 

\section{Background}
\label{sec:background}

\subsection{Reinforcement Learning on robot simulation}
\label{subsec:rl_robotic}
 
RL algorithms have demonstrated promising results in robotics \cite{c12, c13}. However, the acquisition of sufficient training data is challenging and time consuming, especially on real robots. This is why many works in robotics use simulation to train the models before transferring the result to real robots. From a time perspective, this advantage of using simulation rather than a physical prototype is only effective if the simulation runs at least in real time. To measure the temporal performance of a simulator, the ratio between simulation time $\Delta t_{simu}$ and time to perform this simulation $\Delta t_{real}$, called the Real Time Factor $RTF = \frac{\Delta t_{simu}}{\Delta t_{real}}$, is used. The RTF is a unitless quantity used to compare different simulations between them or to the real-time. Real time simulation is possible if $RTF \geq 1$ \cite{c14}. 

The use of FEM in soft-robotics often requires many calculations to simulate a time step. The corresponding models are therefore not compatible with the use of learning algorithms. For example, RL algorithms such as Proximal Policy Optimization (PPO) \cite{c6} used for a locomotion task requires several days of computation \cite{c3} even when model reduction techniques are used: for this robot, considering episodes of 3.6 seconds and a learning of 10000 episodes leads to a simulation time (without learning) of 50 hours.\footnote{All calculations are done on the same machine (macOS Big Sur, Intel Core i7 quad-core 3.1GHz processor, 16GB 2133 MHz LPDDR3 memory), under the same conditions.} For comparison, under the same learning conditions with the low range of rigid robotics simulator RTFs \cite{c14}, it would take 1 hour.\\

\subsection{Inverse modeling and representation space}
\label{subsec:space_and_learning}

There are four spaces to describe the soft robot behavior: the actuation space (e.g. pressure, cable pulling force, torque), the joint space (e.g. cable length or joint angle), the configuration space (e.g. robot deformations) and the task space (e.g. end-effector position for a manipulator, average velocity for locomotors). Solving a TSCC requires linking the task space to the actuation space: this is the role of an inverse model. To obtain this inverse model two solutions are possible: data-based approaches such as RL algorithms \cite{c9}, or physics-based approaches such as FEM-based optimisation approaches \cite{c2}. Data-based approaches require a learning phase that can be long and has to be repeated from scratch for each new system, but allow TSCC to be achieved directly. On the contrary, physics-based approaches allow to obtain a model from the theory but are often difficult to use to solve a TSCC (specifying different points to be reached, defining a sequence of movements to follow, ...). 

It is interesting to consider what information is needed to learn a control strategy in order to model only this information during the learning phase. In the literature, when a physical model is available in motion planning, it is used to link configuration space and actuation space \cite{c23, c24, c25}. Motion planning in configuration space can be left to dedicated algorithms such as RL or RTT algorithms. The general idea is to dissociate the two mapping actuation-configuration spaces and configuration-task spaces. Considering only the second one in the motion planning step allows to deal with redundant actuation and to define higher level tasks during learning.  

\section{Modeling and motion planing using a proxy model}
\label{sec:method}

\subsection{Method overview and setup}
\label{subsec:behavior}

The proxy of a humanoid robot is typically constructed to produce a feasible trajectory in task space, i.e. a trajectory that can be reached by the complete system. This requires to know the robot's workspace, which evaluation has been investigated for several decades in rigid robotics based on kinematic models. The evaluation and characterization of a soft robot's workspace can be far more difficult, due to its dependency to interaction forces and therefore the necessity to use more complex mechanical models. In addition, the potential actuation redundancy will lead to a multiplicity of configurations reaching the same task positions, these configurations being challenging to identify. Therefore, we propose a motion planning method where we define \textit{feasible trajectories} as trajectories that can be approached with some error. Moreover, these trajectories are primarily generated in the robot's configuration space. The method is composed of three steps:
\begin{itemize}
    \item Step 1: Modeling. The proxy is constructed using computational cost efficient mechanical elements to approximate the  deformations of the soft robot and its interactions with the environment.
    \item Step 2: Trajectory generation. The proxy is used to generate feasible trajectories in the configuration space using RL. Although we don't ensure the trajectory can be perfectly followed by the robot, methods can be used to reduce the gap.
    \item Step 3: Transfer. The planned trajectory is used as a reference for an optimization-based inverse control. This control is based on an accurate FEM model, providing actuation inputs that can be reached by the soft robot. This step is used to move from the configuration space to the joint space of the robot.
\end{itemize}
Each step is described more into details in the following.

\subsection{Step 1: Modeling}
\label{subsec:modeling}

A large part of the soft robots proposed so far are made of elements whose behavior can be described by the bending of a neutral fiber. This is the case of locomotive legs, fish robots or trunk type serial manipulators \cite{c1}. We assume that these bending elements can be identified. We can then base the construction of the proxy model on the theory of beams by limiting ourselves to inextensible bending beams. Furthermore, we assume that the actuation imposes a constant pre-curvature on the beams, as conventional actuators such as cable actuators or pneumatic actuators do in free space \cite{c32}. As a consequence, the actuated degrees of freedom can be directly expressed in the configuration space.

The creation of the proxy is done in three steps. In the first step, we identify sub-parts of the soft robot that have similar behaviors in terms of bending and rotation/translation. These sub-parts can be identified by looking at the location and nature of the actuators. In particular, a conventional flexible pneumatic actuator with a cavity to be inflated can be represented by a beam with just one actuated bending. An elastic body deformed by pulling on at least three cables can be modeled using a beam with two degrees of bending actuation.
In the second step, we extract the neutral fiber of the identified sub-parts to create a set of bending beams, called the skeleton of the soft robot. We chose to represent these beams with an Euler-Bernouilli model \cite{c31}

\begin{equation}
    P_{i, t} \sim f(L_b, S_b, M_b, \alpha_{b,t}, P^0_{b, t}, Q^0_{b, t}) , b \in \mathcal{B} 
\label{eq:curvature}
\end{equation}

where the position $P_{i,t}$ points uniformly distributed along the beams $\mathcal{B}$ depends of $\alpha_{b,t}$ the pre-curvature of the beam $b$, $P^0_{b, t}$ the position and $Q^0_{b, t}$ orientation of the base of the beam, $L_b$ the length, $S_b$ the stiffness, $M_b$ the uniform mass. This equation includes design-related parameters ($L_b$, $S_b$, $M_b$) and actuation-related parameters ($\alpha_{b,t}$, $P^0_{b, t}$, $Q^0_{b, t}$). The value of the pre-curvature takes into account the actuation forces applied on the proxy. The actual curvature is given by the value of the pre-curvature and the external forces such as contacts that can deform the beam. The values of $P^0_{b, t}$ and $Q^0_{b, t}$ are either passive or actuated. The points $P_{i,t}$ represent the configuration of the soft robot. Considering beams' positions and not beams' curvature for the robot's state allows to model more efficiently the interactions with the environment such as contacts, amongst other advantages \cite{c23}. To avoid numerical singularity, the beams are constructed such that their extremities either connect to another beam, connect to the environment (like the tip of a locomotor's leg) or are controlled (like the tip of a manipulator). In the case where there are free beam ends, additional non-actuated beams are added, always remaining consistent with the geometry of the robot. All these beams allow to take into account the compliance of the robot structure. 

Finally in a third step, a surfacic mesh representing the external surface of the soft robot is added on top of this assembled skeleton. This mesh is mapped onto the robot skeleton through barycentric mappings which create a link between the surface contact forces and the resultant forces distributed on the beams \cite{c17}. It is used to take into account the interactions between the robot and its environment.

The proxy creation requires the choice of parameters (maximum curvature $\alpha_{b}^{max}$, mass $M_b$, stiffness $S_b$). In this paper, we use a Bayesian approach \cite{c15} with a few iterations to optimize them. To obtain quick realistic results, we restrict the search space around hand-found values that give a realistic behavior in terms of deformation. As the proxy model is a simplified model, it is not useful to obtain the "best" parameters to describe it at the risk of spending a lot of time for minimal transfer results. The cost function used for the Bayesian optimization is given by the $L_2$ distance between points mapped on the faithful and proxy models. Both models follow an identical trajectory in terms of deformation and must exhibit similar behaviors on this trajectory. The trajectory is separated into optimization and validation positions, which enables to check that the proxy is interpolating, i.e. that the parameters found for the optimization positions are also good for the validation positions.

We have a lot of freedom to create the proxy. Other mechanical elements could be used for approximating the robot behavior as discussed in section \ref{subsec:discussion_modeling}, and different construction methods for the skeleton could be envisioned. The idea here is to demonstrate that we can use a very simplified proxy of the robot for control but not to demonstrate that our choice of proxy model is optimal.

\subsection{Step 2: Trajectory generation}
\label{subsec:learning}

Step 1 provides a proxy that represents the robot in its configuration space. Using this model, and thus the definition of the robot in this space, a configuration sequence can be found to solve a given task. In this paper, we use a RL algorithm to learn this configuration sequence, called the strategy. RL allows to solve TSCC and to adapt in case of perturbations during these tasks. The trajectory is not generated in one single block but sequentially, one simulation step after the other. In the configuration space, the actions are the global deformation represented by the beams' pre-curvature $\alpha_{b,t}$ in eq. \ref{eq:curvature}, and the eventual translations $P^0_{b, t}$ and  $Q^0_{b, t}$ rotations of the beams extremities. The points of beams $P_{i,t}$ in eq. \ref{eq:curvature} uniformly distributed on the proxy model naturally represent the RL state of the proxy.  

RL algorithms tend to exploit the weaknesses of the underlying model to find an optimal strategy. Since our proxy model is an approximate model, there is no guarantee that the strategy found can be exactly realized by the real robot (in terms of deformation and contact with the environment), although the proxy's parameters have been optimized to match the faithful model. In order to overcome this problem at the learning step, it is possible to introduce information about the a-priori behavior of the robot, e.g. symmetry of actuation or limitation of the actuation amplitude. This can be done by reducing the search space of the learned strategy and depends on the task at hand. We will illustrate this in sections \ref{sec:experiment} and \ref{sec:results}.

\begin{figure}[!ht]
\centering
\resizebox{.5\textwidth}{!}{
\includegraphics{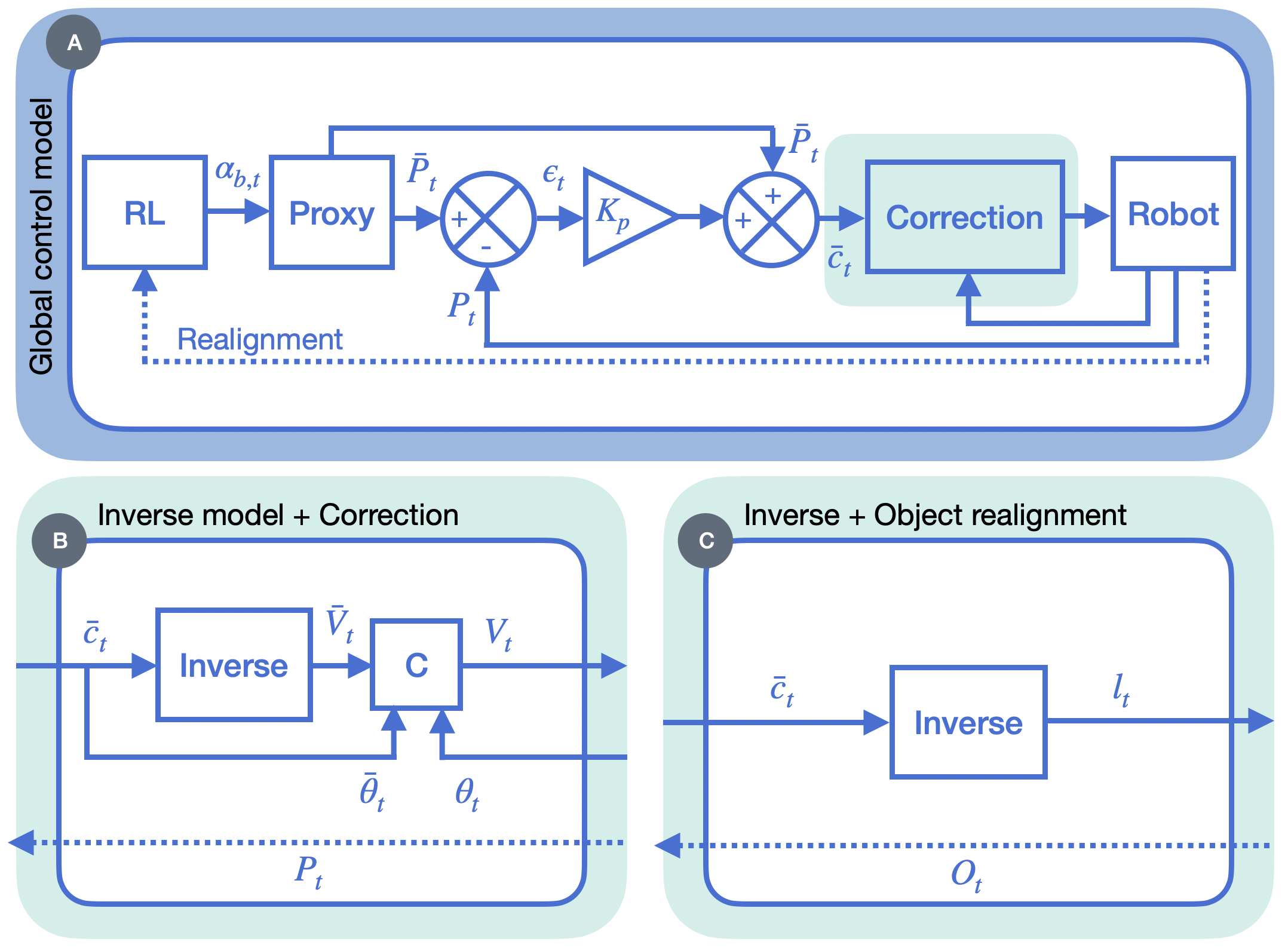}
}
\caption{Control scheme used during the transfer. (A) General case. A trained RL algorithm gives a curvature instruction $\alpha_{b,t}$ to the proxy model to solve a task. Following this instruction, a position reference $\bar P_t$ is extracted and compared to the current position of the robot $P_t$ to form a corrected position command $\bar c_t$. The realignment allows taking into account the position actually reached by the robot $P_t$ or the manipulated objects $O_t$. An inverse model gives an actuation command that can be corrected by taking into account information from both the proxy model and the robot. (B) The inverse model is used to find a pneumatic actuation $V_t$. A corrector $C$ is used to correct the actuation to take into account the angles observed in the proxy $\bar \theta_t$. (C) Only the inverse model is used to find the cable lengths $l_t$.  }
\label{im:automatic}
\end{figure}

\subsection{Step 3: Transfer}
\label{subsec:transfer}

Step 2 gives a configuration trajectory of the proxy that solves the desired task. The last step consists in transferring this trajectory to the robot joint space using the high-fidelity FEM model. We implement a closed-loop position control based on the optimization-based inverse model in~\cite{c2} and follow the feasible trajectory (in the sense of the feasibility defined in Section \ref{subsec:modeling}) in simulation. In addition, two other control loops are used (see Figure \ref{im:automatic}). The first one corresponds to an external control loop linking the RL controler to the complete robot model, as can be seen in Figure \ref{im:automatic}.A. Due to friction contacts, some accumulation of errors can lead to a difference in behavior between the complete model and the proxy model, and a step of realignement of the proxy model is necessary. To do this, relevant positions in the complete model are retrieved and imposed on the proxy robot model. It enables to obtain a proxy instruction that adapts to what happens with the complete robot model. The second control loop partially compensates for certain local differences that may exist between the proxy model and the complete model by correcting the actuation given by the inverse model. Depending on the application (locomotion, manipulation), these differences will involve different elements such as contact points or objects to be manipulated. In our method, realignement imposes the position of the object handled by the complete model on the object handled by the proxy model. Figure 2.B and 2.C represent the correction modules in Figure 2.A for locomotion and manipulation respectively.

\section{Experiment}
\label{sec:experiment}

\subsection{Application cases: robots and tasks}
\label{subsec:robots}

\begin{figure}[!ht]
\centering
\resizebox{.5\textwidth}{!}{
\includegraphics{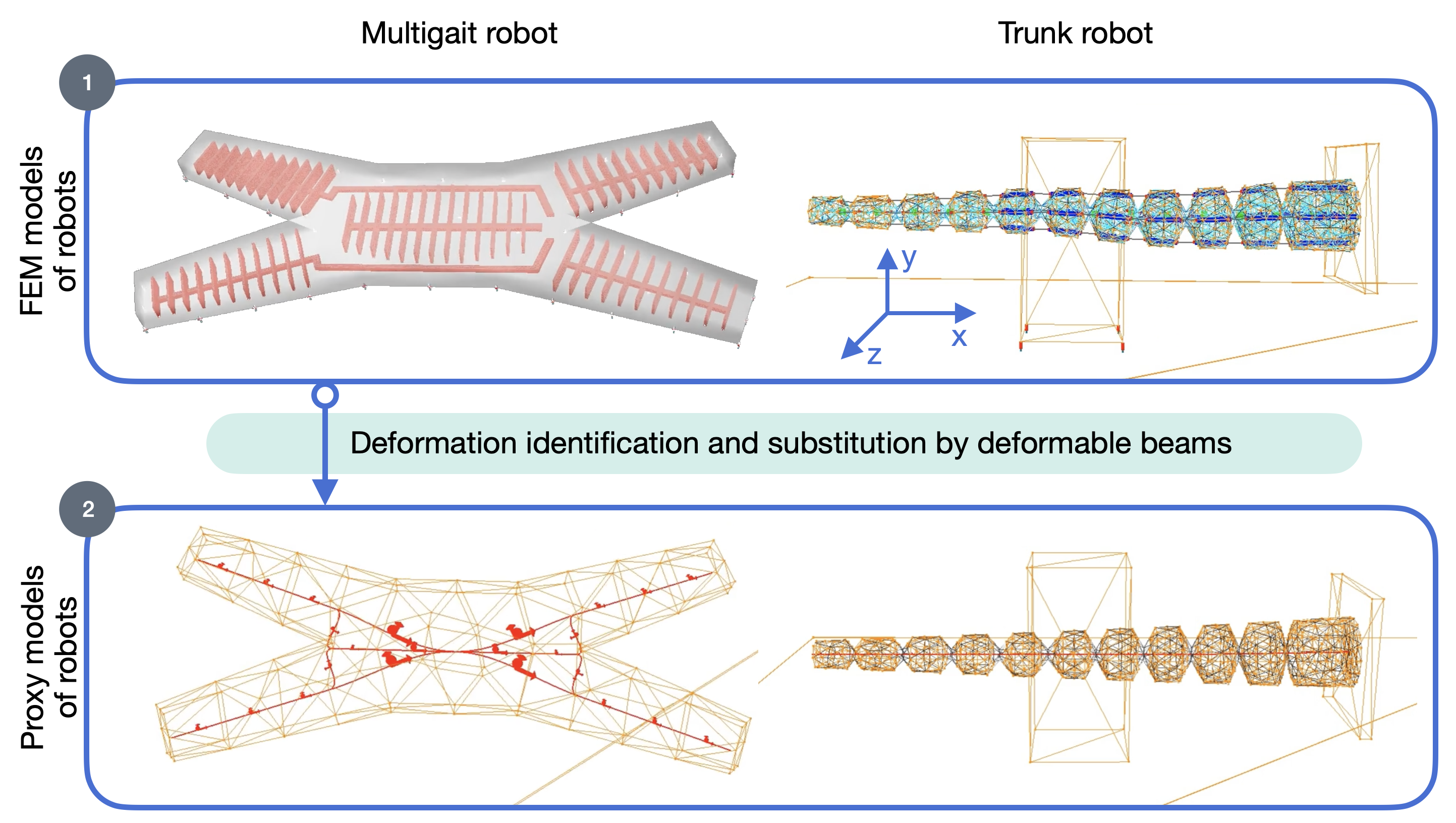}
}
\caption{Representation of the FEM (1) and proxy models (2) of the Multigait and Trunk soft robots. The deformable beams composing the skeleton of the proxy model are depicted in red. The external meshes for both models are identical.}
\label{im:models}
\end{figure}

For the purpose of this paper, we consider two robots designed to perform locomotion and manipulation tasks respectively. These two robots are represented in Figure \ref{im:models}.1. 

The first robot presented in \cite{c8} is used for locomotion tasks. In the rest of this article we will refer to it as the Multigait robot. It is composed of four legs and a central part, made of silicone. Each of these elements has one cavity. The pressure inside the cavities can be changed and the higher the pressure the more the leg bends. The objective of this robot is to move forward as fast as possible. The Multigait is built according to geometric symmetry planes. Control strategies can exploit these symmetries to achieve particular movements: for example, the symmetry of the left and right legs can be used to move forward, which leads to the same actuation of the front legs (resp. the back legs). This robot is a well known soft robot in the community. Even with model reduction methods \cite{c16}, its simulation is too slow to be applied with RL techniques (RTF less than 0.1). It also has the characteristic of not being slender, which makes proxy modeling less intuitive.

The second robot presented in \cite{c2} is used for manipulation tasks. In the rest of this article we will refer to it as the Trunk robot. It is composed of two segments that can be actuated in bending using two pairs of antagonistic cables per segment each. In addition, its base can translate in the $x$ direction. The environment of the Trunk robot is composed of a rigid cube that can glide on a rigid planar surface. The objective of this robot is to bring the center of mass of the cube to a certain position of the plane $(x,z)$. The Trunk robot has a large action space (8 cables and 1 translation movement) with redundant actuation (more actuators than actuated DoFs). The interaction with an object also makes learning more time consuming.

These two robots are simulated using the Simulation Open Framework Architecture (SOFA) software \cite{c17} , and in particular the plugin soft robot which integrates numerous modeling and control tools for soft robots and their environment. It also allows the simulation of particular structures such as beams and truss \cite{c18}. The two robots were already modeled using SOFA and their simulations are considered as ground truth since the transfer between simulation and real robot has already been done in previous works \cite{c2, c16}.

\subsection{Proxy modeling of Multigait and Trunk robots }
\label{subsec:proxies}

Following the steps described in section \ref{subsec:modeling}, the two robots' proxies are shown in Figure 3.2.

The Multigait robot is a robot with 5 cavities and each cavity represents a bending degree of freedom. For the proxy, we therefore take the robot skeleton and replace each cavity with a beam with one bending degree of freedom. Additional beams are added following the edges of the robot in order to satisfy the condition on beam coupling introduced in section \ref{subsec:behavior}. 

The Trunk robot can be seen as two beams placed end to end and each having two degrees of freedom of bending. The trunk is thus separated into two parts, and each of these parts is represented by a beam. The mobile base of the robot, which can translate along the $x$-axis, is the same for the proxy model and the complete model. 

\subsection{Solving the task using RL algorithm on the proxy model}
\label{subsec:RLproxy}

For the generation of the trajectory with RL, we use SofaGym \cite{c3} for the implementation of the environment and the PP0 algorithm \cite{c6} with the following characteristics. We emphasize that the contribution of this work is independent of the chosen RL algorithm.

For the Multigait robot, the state of the robot is defined by the 20 points uniformly distributed on its skeleton. We limit the action space by considering the all-or-nothing positive bending of actuators, while imposing symmetric activation to go forward. Since we have an episode with a fixed duration, maximizing the speed is equivalent to maximizing the distance covered during this duration. The reward is defined as the position increment at each time step. For this environment, one iteration in SofaGym corresponds to 0.6s of simulation. For the Trunk robot, the state is defined by the 16 points uniformly distributed on its skeleton, the position of the center of the cube and the position of the goal. The actions are defined by the 2 possible bendings of the 2 beams and the translation of its base along its main axis. The main reward is given by the distance between the cube and the goal. We also added a negative weighted sum of the distance between the trunk and the cube to favor the contact. For this environment, one iteration in SofaGym corresponds to 0.3s of simulation. The task is solved when the cube is less than 8mm from the goal which corresponds to 5\% of the trunk length.

\section{Results}
\label{sec:results}

Once the proxies are defined, we have to perform steps 2 and 3 of the method, namely the trajectory generation and the transfer to the complete model. We present in this section the results obtained in terms of simulation time, parameter optimization, learning, and transfer. 

\subsection{Improved simulation time and learning time}

The simulation times are evaluated in terms of RTF. All tests are performed under the same calculation conditions. The proxy and FEM models are put in the configuration where the number of contacts is maximal. The different results are summarized in Table \ref{tab:time}. The proxy is 250 times faster for the Multigait and 3 times faster for the Trunk compared to the FEM model. The RTF gain is much larger for the Multigait than the Trunk due to the difference in FEM mesh sizes, which are 34k nodes and 900 nodes respectively. Moreover, in the context of this study, the proxy models can have $RTF>1$ that corresponds to a computation time compatible with real-time, which is advantageous in a learning framework. The calculation is as fast as retrieving the data for live learning on a real robot. The method is frugal, in the sense that it is less computationally expensive than the use of the complete models. Compared to the other approach consisting of solving heavy models in parallel, the gain in computation time is obtained here without increasing the computation burden.

\begin{table}[!h]
\centering
   \caption{RTF obtained for the two robots, Mu. = Multigait, Tr. = Trunk, F. = FEM, P. = Proxy, MOR = Model Order Reduction}
    \centering
    \begin{tabular}{| c | c | c | c | c| c |}
        \hline
         $\Delta t_{simu}$ & F. Mu. & MOR Mu. & P. Mu. & F. Tr. & P. Tr.\\ \hline
         
         0.01 & 0.002 & 0.06 & 0.57 & 0.22 & 0.62 \\ \hline
         
         0.02 & 0.004 & 0.12 & 1.06 & 0.49 & 1.33 \\ \hline
         
         0.03 & 0.006 & 0.17 & 1.51 & 0.72 & 1.96 \\ \hline
        
    \end{tabular}

    \label{tab:time}
\end{table}

\begin{figure*}[!ht]
\centering
\resizebox{0.92\textwidth}{!}{
\includegraphics{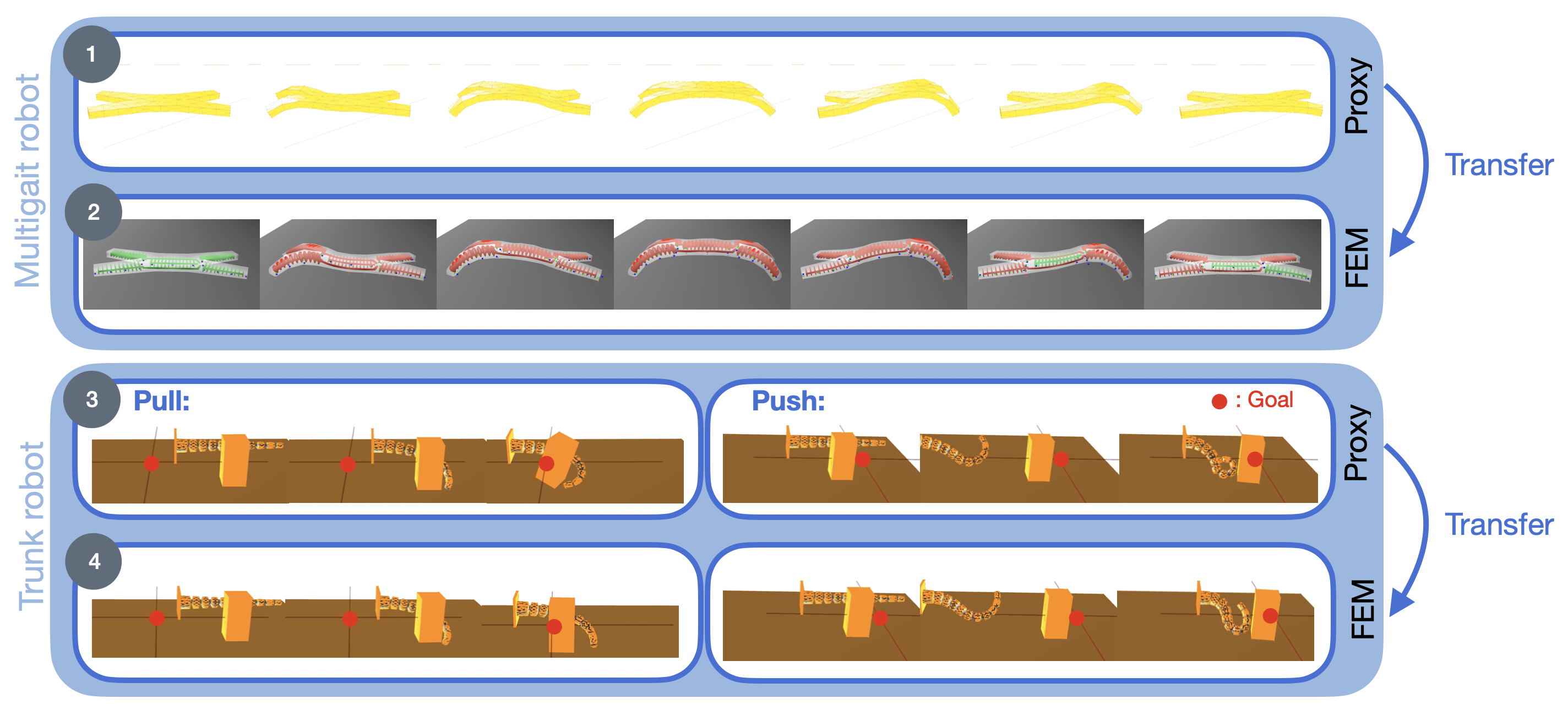}
}
\caption{Example of strategy obtained on the proxy model and transferred to the complete model in the case of Multigait and Trunk robots. (1) Strategy learned with the proxy model of the Multigait robot. (2) Transfer of the strategy to the FEM model of the Multigait robot. (3) Strategy learned with the proxy model of the Trunk robot. (4) Transfer of the strategy to the FEM model of the Trunk robot. Despite differences in deformation, the strategy can be transferred between these two models. }
\label{im:transfert}
\end{figure*}

\subsection{Proxies parameters optimization}
\label{subsec:opti}

The actions used to generate the parameter optimization trajectory are actions that bring the robot into extreme bending and resting configurations. We fix the time step to 0.01s, and we split the trajectories in optimisation en validation sets. For the Multigait robot, we consider a 10.8s trajectory composed of all-or-nothing actions. We use the position of the bottom face in the two models for the optimization. For the Trunk robot, we use the axial symmetry. We consider a 0.6s trajectory composed of 2 actions. The positions of the external mesh points are used for optimization. As the optimization was done without interaction between the cube and the robot, we had to optimize the cube weight a posteriori to obtain the same behavior during interaction. We could improve that in future work. To ease the transfer (described in \ref{subsec:transferMultiAndTrunk}), a safety parameter is used to force the proxy to find underestimated actuation magnitudes compared to the actuation limits of the actual system. This allows compensating for the deformation error between the models. 

For the Multigait robot the average error defined in \ref{subsec:modeling} on the optimization positions is 8.02 mm (4\% of the characteristic size of the robot), and 10.98 mm for the intermediate positions (5.5\% of the characteristic size of the robot). The characteristic size of the robot corresponds to its length along the axis of motion, i.e. along the $x$ axis. For the Trunk robot, we obtain with the optimization an average error of 10.2 mm on the positions used for the optimization and 9.85 mm on the intermediate positions. This error corresponds to about 5.2\% of the robot's length.
% characteristic length

\subsection{Task resolution and transfer to the complete model}
\label{subsec:transferMultiAndTrunk}

For both robots, the PPO algorithm manages to find a strategy on the proxy that solves the task. These strategies are shown in Figure \ref{im:transfert}. The strategy transfer for the Multigait robot and for the Trunk robot are illustrated in Figure \ref{im:transfert}.2 and \ref{im:transfert}.4 respectively. We consider that the transfer is successful when the models have similar behaviors with respect to the cost function used for the training.

For the Multigait robot, we use the actuation correction shown in Figure \ref{im:automatic}.B. The inverse model is used to calculate the actuation according to the position reference given by the proxy model. Since the inverse model does not take into account the angle of incidence of the contact points, a correction step allows to follow both the position and orientation of the contact points. The steps of realignement are done with a fixed time step. After the transfer, the complete model advances by about 5.25 mm against 5.52 mm for the proxy model, corresponding to a difference of~5\%. 

For the Trunk robot, we study two cases: the positioning of the cube along the $x$-axis and the positioning of the cube at a position $(x,z)$. The used controller is shown in Figure \ref{im:automatic}.C and is simpler than the one used for the Multigait robot. The registration step is used to take into account the position of the cube: after several actions, the position of the cube detected in the complete model is imposed on the cube of the proxy model. Thanks to this, if the cube is not in the expected place, there is a re-planning step where the RL-based motion planning adapts the actuation of the trunk in order to ensure the success of the manipulation task. This re-planing step can be seen in the associated video. For the positioning along the $x$-axis, the relative positioning error of the cube $\frac{|r_t^{proxy}-r_t^{FEM}|}{|r_t^{FEM}|}|$, where $r_t$ corresponds to the reward explained in \ref{subsec:RLproxy}, is about 5\% between the two models. We have the same results when the goal is in a position $(x,y)$, even if the position is outside the learning space. In the latter case, the final position of the cube corresponds to an error of 7\% compared to the position given by the proxy model. 

\subsection{Trajectory analysis and proxy accuracy}
\label{subsec:TrajectoryLearningAnalysis}

\begin{figure}[!h]
\centering
\resizebox{0.48\textwidth}{!}{
\includegraphics{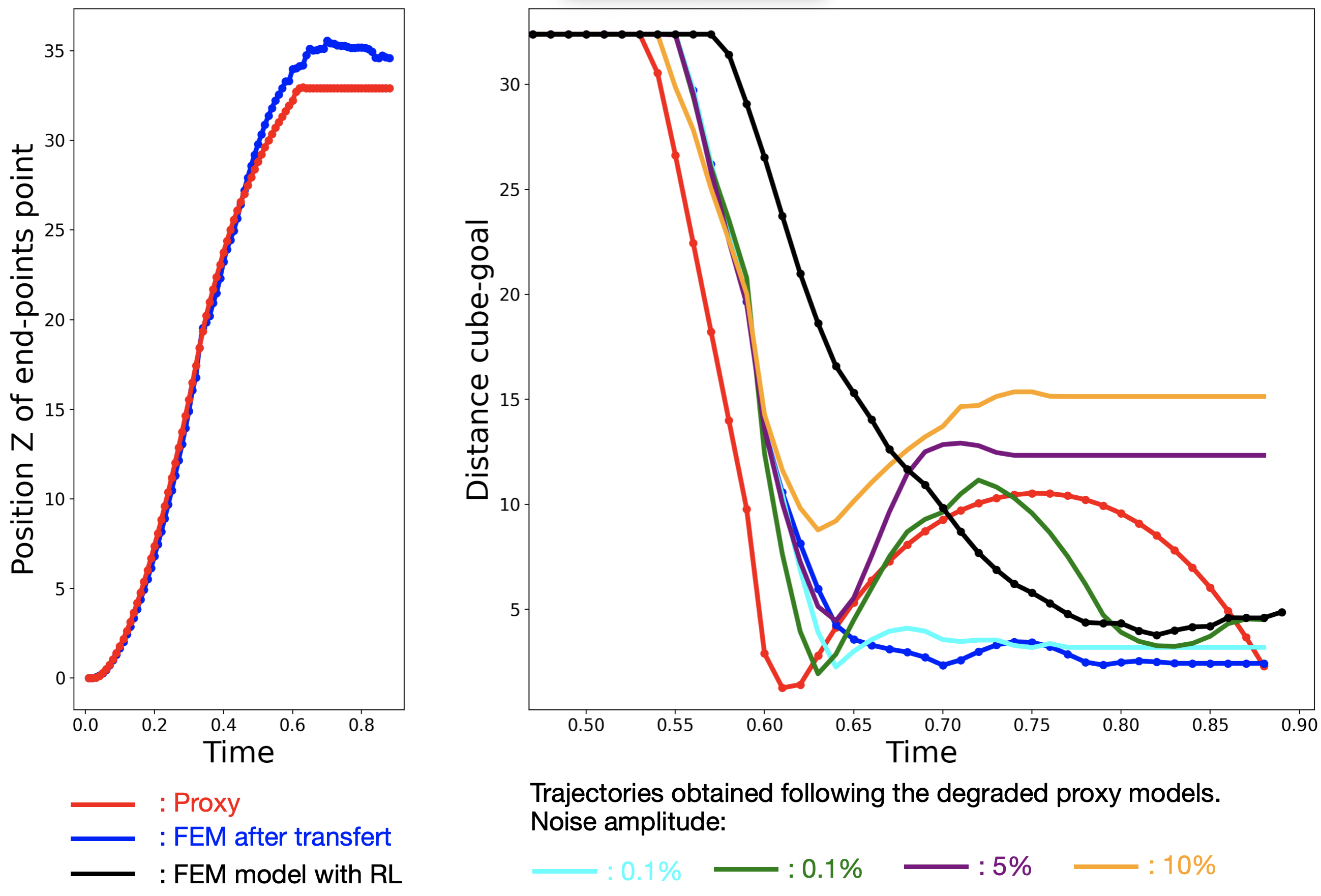}
}
\caption{Results obtained for the Trunk robots. On the left: Position of the tip of the Trunk over time for the proxy (red) and the FEM (blue) after transfer. On the right: Distance between the cube and the goal. In red: Reference proxy model. In blue: FEM model after transfer. In black: FEM model with a RL algorithm (without proxy). Other colors: FEM model following the degraded proxy models, with 0.1\% (cyan), 1\% (green), 5\% (purple), and 10\% (orange) of noise amplitude.}
\label{im:results_trunk}
\end{figure}

We analyze further the results of our approach in the case of the Trunk robot. Figure \ref{im:results_trunk} left shows the trajectory of the robot tip along the Z-axis obtained with the proxy and the complete model. The two models are close in terms of positioning at the point of contact between the cube and the Trunk. However, the behavior is different for the tip where the error increases with the robot deformation. In addition, we can see on the right that there is a bell in the distance measured using the proxy model (red line) and not in the FEM model (blue line). This is due to the absence of friction between the cube and the ground in the proxy. However, the global behavior is the same, with the distance being minimized with both models. For the Multigait robot, we obtain a worse match compared to the Trunk given the less adequate hypothesis of a skeleton based model for the proxy. To improve the transfer, we found that it is key to control the contact points, more than matching the deformations.

We also analyze the impact of proxy accuracy by creating degraded proxies, less accurate, and by considering the use of RL on the complete FEM model directly. Degraded proxies are obtained by adding uniform noise to the reference proxy parameters of magnitude 0.1-10\%. The strategy is learned with the degraded proxy and the transfer is performed with the nominal FEM model. The results are presented in Figure \ref{im:results_trunk}. We can see that in each case our planning method leads to a trajectory that decreases the distance between the cube and goal positions, demonstrating a certain robustness to the proxy accuracy. Similar results are obtained when the parameters are close to those of the reference proxy (green or cyan lines). However, poorly chosen parameters can lead to degraded results, whether in terms of the reward of the proxy model or transfer (orange or purple lines). Concerning learning with the complete model, learning stops when the distance between the cube and the goal is less than 5 mm, and both the FEM model with RL and the proxy method manage to solve the task. It seems that, in this particular case of pushing the cube, the proxy leads to better results (2 mm between the cube and the goal for the proxy method, against 4.5 mm for the FEM model with RL). In any case, the task is solved using a proxy model with a highly reduced computation time, which shows its interest.

\section{Discussion}
\label{sec:discussion}

The previous results show that the method presented in section \ref{sec:method} allows to control a robot in locomotion and manipulation tasks. We discuss here its different steps.

\subsection{Discussion on the proxy modeling step}
\label{subsec:discussion_modeling}

The movements of the proxies developed in this study are parameterized using their deformation, in particular curvature parameters, and not directly with their actuation. This allows having a minimum set of geometric parameters for the motion planning, in particular in case of redundant actuation (like in the trunk case). Focusing on the deformations allows for decreasing the dimension of the action space during the learning phase and moving the management of antagonistic actuation to the transfer step. As a consequence, the temporal performances of the simulations and the motion planning in general are improved. Although we have considered inextensible beams, taking into account extension and even torsion is theoretically possible, for example with Cosserat models of the beam \cite{c18}. In practice, we didn't find any problem for controlling the Trunk, but this question remains in other cases with different interactions with the environment or with a different robot. Finally, the use of beam elements is one of many approximations that could be used to build a proxy model. It would be possible to use other approximations such as mass-spring, articulated bodies, or particles to construct the proxy. We could also use reduced-order models. The complete method would remain the same, performing the training on a simplified model then transferring it on a complete model. 

\subsection{Discussion on the trajectory generation step}
\label{subsec:discussion_motion}

In our work, the focus was not on the sophisticated RL strategy. The strategies found are for simple tasks, and we were satisfied with trajectories that solve the task. Since we know that the proxy model is an approximate model, the optimal solution found in the configuration space is not necessarily optimal for the full model. The strategy, however, solves the task even with the complete model, and we have focused on the general method and transfer rather than on obtaining the optimal strategy with the proxy for a wide variety of tasks. For the Trunk robot, using RL on several goals allows for adapting the instruction after a realignment step. For the Multigait robot, we use the symmetry defined in the section \ref{subsec:robots} to constrain the strategy. The result is a less complicated strategy than with non-symmetrical continuous actuation, but easier to transfer.  

\subsection{Discussion on the transfer step}
\label{subsec:discussion_transfer}

One of the key feature of the method is to simulate the transfer. Compared to the methods directly realizing the passage between the simplified model (particle, voxel, ...) and physical prototype, we have an intermediate step. It is possible to test and correct the learned strategy using the FEM model before transferring it to a physical prototype and see if the strategy is viable or not. Generating the actuation trajectory with the inverse method on the full FEM model allows to correct the errors of the proxy and increasing the chances of successful transfer. Using FEM simulation allows to rely on models whose transfer has already been validated by numerous publications, including a wide variety of soft robotics systems such as manipulators, grippers, and mobile robots. We are convinced that if we manage to transfer the policies learned on a FEM model, there is a good chance that they will be transferred to a physical prototype. The most important point is therefore the transfer to the FEM model, the transfer to a physical prototype being another topic we will investigate in the future.

\section{Conclusion and future work}
\label{conclusion}

In this paper, we propose a motion planning method that can be applied to complex soft robotics systems and successfully solves manipulation and locomotion tasks. It relies on the use of a faithful FEM model of the system and a simplified and frugal proxy model composed of deformable beams. After parameter optimization, the position error of the proxy model is about 5\% of the robots' characteristic length. Moreover, real-time computation is achieved, making it suitable for learning strategies. Trajectories are then generated using RL and correction steps, which leads to successfully performing the tasks while leveraging the soft robots' symmetries and actuation redundancy. Finally, the learned strategies are transferred to the faithful FEM model to generate the final trajectory in actuation space. The implementation of this transfer in simulation allows in particular for correcting the errors introduced by the simplified model. As discussed in section \ref{sec:discussion}, the modeling, learning, or transfer steps can be improved to facilitate transferable learning from a proxy model and bring its optimal trajectory closer to the complete model's one.

\addtolength{\textheight}{-12cm}   % This command serves to balance the column lengths

%%%%%%%%%%%%%%%%%%%%%%%%%%%%%%%%%%%%%%%%%%%%%%%%%%%%%%%%%%%%%%%%%%%%%%%%%%%%%%%%

%%%%%%%%%%%%%%%%%%%%%%%%%%%%%%%%%%%%%%%%%%%%%%%%%%%%%%%%%%%%%%%%%%%%%%%%%%%%%%%%

%%%%%%%%%%%%%%%%%%%%%%%%%%%%%%%%%%%%%%%%%%

%%%%%%%%%%%%%%%%%%%%%%%%%%%%%%%%%%%%%%%%%%%%%%%%%%%%%%%%%%%%%%%%%%%%%%%%%%%%%%%%

\bibliography{bibliography}

\begin{thebibliography}{10}
\providecommand{\url}[1]{#1}
\csname url@samestyle\endcsname
\providecommand{\newblock}{\relax}
\providecommand{\bibinfo}[2]{#2}
\providecommand{\BIBentrySTDinterwordspacing}{\spaceskip=0pt\relax}
\providecommand{\BIBentryALTinterwordstretchfactor}{4}
\providecommand{\BIBentryALTinterwordspacing}{\spaceskip=\fontdimen2\font plus
\BIBentryALTinterwordstretchfactor\fontdimen3\font minus \fontdimen4\font\relax}
\providecommand{\BIBforeignlanguage}[2]{{%
\expandafter\ifx\csname l@#1\endcsname\relax
\typeout{** WARNING: IEEEtran.bst: No hyphenation pattern has been}%
\typeout{** loaded for the language `#1'. Using the pattern for}%
\typeout{** the default language instead.}%
\else
\language=\csname l@#1\endcsname
\fi
#2}}
\providecommand{\BIBdecl}{\relax}
\BIBdecl

\bibitem{c1}
B.~Mazzolai and M.~Cianchetti, ``Soft robotics: Technologies and systems pushing the boundaries of robot abilities,'' \emph{Science Robotics}, vol.~1, 2016.

\bibitem{c2}
E.~Coevoet, A.~Escande, and C.~Duriez, ``{Soft robots locomotion and manipulation control using FEM simulation and quadratic programming},'' in \emph{{RoboSoft 2019 - IEEE International Conference on Soft Robotics}}, Seoul, South Korea, Apr. 2019.

\bibitem{c3}
E.~M{\'e}nager, P.~Schegg, E.~Khairallah, D.~Marchal, J.~Dequidt, P.~Preux, and C.~Duriez, ``{SofaGym: An open platform for Reinforcement Learning based on Soft Robot simulations},'' \emph{{Soft Robotics}}, 2022.

\bibitem{c9}
S.~Bhagat, H.~Banerjee, Z.~T.~H. Tse, and H.~Ren, ``Deep reinforcement learning for soft, flexible robots: Brief review with impending challenges,'' \emph{Robotics}, vol.~8, p.~4, 2019.

\bibitem{c4}
B.~Homberg, R.~K. Katzschmann, M.~R. Dogar, and D.~Rus, ``Robust proprioceptive grasping with a soft robot hand,'' \emph{Autonomous Robots}, vol.~43, pp. 681--696, 2018.

\bibitem{c5}
H.~Wang, J.~Chen, H.~Y.~K. Lau, and H.~Ren, ``Motion planning based on learning from demonstration for multiple-segment flexible soft robots actuated by electroactive polymers,'' \emph{IEEE Robotics and Automation Letters}, vol.~1, pp. 391--398, 2016.

\bibitem{c21}
X.~Wang, Y.~Li, and K.~W. Kwok, ``A survey for machine learning-based control of continuum robots,'' \emph{Frontiers in Robotics and AI}, vol.~8, 2021.

\bibitem{c2a}
H.~Choi, C.~Crump, C.~Duriez, A.~Elmquist, G.~D. Hager, D.~Han, F.~Hearl, J.~K. Hodgins, A.~Jain, F.~A. Leve, C.~Li, F.~Meier, D.~Negrut, L.~Righetti, A.~Rodriguez, J.~Tan, and J.~C. Trinkle, ``On the use of simulation in robotics: Opportunities, challenges, and suggestions for moving forward,'' \emph{Proceedings of the National Academy of Sciences of the United States of America}, vol. 118, 2020.

\bibitem{c2b}
M.~Graule, T.~McCarthy, C.~Teeple, J.~Werfel, and R.~Wood, ``Somogym: A toolkit for developing and evaluating controllers and reinforcement learning algorithms for soft robots,'' \emph{IEEE Robotics and Automation Letters}, vol.~7, pp. 1--1, 04 2022.

\bibitem{c19}
Y.~Hu, J.~Liu, A.~E. Spielberg, J.~B. Tenenbaum, W.~T. Freeman, J.~Wu, D.~Rus, and W.~Matusik, ``Chainqueen: A real-time differentiable physical simulator for soft robotics,'' \emph{2019 International Conference on Robotics and Automation (ICRA)}, pp. 6265--6271, 2018.

\bibitem{c20}
Y.~Hu, L.~Anderson, T.-M. Li, Q.~Sun, N.~A. Carr, J.~Ragan-Kelley, and F.~Durand, ``Difftaichi: Differentiable programming for physical simulation,'' \emph{ArXiv}, vol. abs/1910.00935, 2019.

\bibitem{c22}
I.~A. Seleem, H.~El-Hussieny, S.~F.~M. Assal, and H.~Ishii, ``Development and stability analysis of an imitation learning-based pose planning approach for multi-section continuum robot,'' \emph{IEEE Access}, vol.~8, pp. 99\,366--99\,379, 2020.

\bibitem{c23}
B.~H. Meng, I.~S. Godage, and I.~Kanj, ``Rrt*-based path planning for continuum arms,'' p. 6830–6837, Jul. 2022.

\bibitem{c24}
Z.~Hawks, C.~G. Frazelle, K.~E. Green, and I.~D. Walker, ``Motion planning for a continuum robotic mobile lamp: Defining and navigating the configuration space,'' \emph{2019 IEEE/RSJ International Conference on Intelligent Robots and Systems (IROS)}, pp. 2559--2566, 2019.

\bibitem{c25}
D.~Braganza, D.~M. Dawson, I.~D. Walker, and N.~Nath, ``A neural network controller for continuum robots,'' \emph{IEEE Transactions on Robotics}, vol.~23, pp. 1270--1277, 2007.

\bibitem{c16}
O.~Goury and C.~Duriez, ``Fast, generic, and reliable control and simulation of soft robots using model order reduction,'' \emph{IEEE Transactions on Robotics}, vol.~34, pp. 1565--1576, 2018.

\bibitem{c26}
D.~E. Orin, A.~Goswami, and S.-H. Lee, ``Centroidal dynamics of a humanoid robot,'' \emph{Autonomous Robots}, vol.~35, pp. 161--176, 2013.

\bibitem{c28}
C.~Mastalli, M.~Focchi, I.~Havoutis, A.~Radulescu, S.~Calinon, J.~Buchli, D.~G. Caldwell, and C.~Semini, ``Trajectory and foothold optimization using low-dimensional models for rough terrain locomotion,'' \emph{2017 IEEE International Conference on Robotics and Automation (ICRA)}, pp. 1096--1103, 2017.

\bibitem{c11}
R.~Budhiraja, J.~Carpentier, C.~Mastalli, and N.~Mansard, ``Differential dynamic programming for multi-phase rigid contact dynamics,'' \emph{2018 IEEE-RAS 18th International Conference on Humanoid Robots (Humanoids)}, pp. 1--9, 2018.

\bibitem{c12}
R.~Liu, F.~Nageotte, P.~Zanne, M.~de~Mathelin, and B.~Dresp, ``Deep reinforcement learning for the control of robotic manipulation: A focussed mini-review,'' \emph{ArXiv}, vol. abs/2102.04148, 2021.

\bibitem{c13}
T.~Haarnoja, A.~Zhou, S.~Ha, J.~Tan, G.~Tucker, and S.~Levine, ``Learning to walk via deep reinforcement learning,'' \emph{ArXiv}, vol. abs/1812.11103, 2018.

\bibitem{c14}
\BIBentryALTinterwordspacing
M.~K{\"{o}}rber, J.~Lange, S.~Rediske, S.~Steinmann, and R.~Gl{\"{u}}ck, ``Comparing popular simulation environments in the scope of robotics and reinforcement learning,'' \emph{CoRR}, vol. abs/2103.04616, 2021. [Online]. Available: \url{https://arxiv.org/abs/2103.04616}
\BIBentrySTDinterwordspacing

\bibitem{c6}
J.~Schulman, F.~Wolski, P.~Dhariwal, A.~Radford, and O.~Klimov, ``Proximal policy optimization algorithms,'' \emph{ArXiv}, vol. abs/1707.06347, 2017.

\bibitem{c32}
I.~Webster, Robert~J. and B.~A. Jones, ``\BIBforeignlanguage{en}{Design and kinematic modeling of constant curvature continuum robots: A review},'' p. 1661–1683, Jun. 2010.

\bibitem{c31}
Y.~Shapiro, K.~G{\'a}bor, and A.~Wolf, ``Modeling a hyperflexible planar bending actuator as an inextensible euler–bernoulli beam for use in flexible robots,'' \emph{Soft robotics}, vol.~2, pp. 71--79, 2015.

\bibitem{c17}
F.~Faure, C.~Duriez, H.~Delingette, J.~Allard, B.~Gilles, S.~Marchesseau, H.~Talbot, H.~Courtecuisse, G.~Bousquet, I.~Peterl{\'i}k, and S.~Cotin, ``Sofa: A multi-model framework for interactive physical simulation,'' 2012.

\bibitem{c15}
B.~Shahriari, K.~Swersky, Z.~Wang, R.~P. Adams, and N.~de~Freitas, ``Taking the human out of the loop: A review of bayesian optimization,'' \emph{Proceedings of the IEEE}, vol. 104, pp. 148--175, 2016.

\bibitem{c8}
R.~F. Shepherd, F.~Ilievski, W.~Choi, S.~A. Morin, A.~A. Stokes, A.~D. Mazzeo, X.~Chen, M.~Wang, and G.~M. Whitesides, ``Multigait soft robot,'' \emph{Proceedings of the National Academy of Sciences}, vol. 108, pp. 20\,400 -- 20\,403, 2011.

\bibitem{c18}
Y.~Adagolodjo, F.~Renda, and C.~Duriez, ``Coupling numerical deformable models in global and reduced coordinates for the simulation of the direct and the inverse kinematics of soft robots,'' \emph{IEEE Robotics and Automation Letters}, vol.~6, pp. 3910--3917, 2021.

\end{thebibliography}
\bibliographystyle{IEEEtran}

\end{document}